\documentclass[a4paper,conference]{IEEEtran}
\usepackage{cite}

\ifCLASSINFOpdf
  \usepackage[pdftex]{graphicx}
\else
\fi
\usepackage{amsmath}
\usepackage{algorithmic}

\usepackage{array}

\ifCLASSOPTIONcompsoc
  \usepackage[caption=false,font=normalsize,labelfont=sf,textfont=sf]{subfig}
\else
  \usepackage[caption=false,font=footnotesize]{subfig}
\fi
\usepackage{url}

\graphicspath{{figures/}}
\usepackage{xcolor}
\usepackage{multirow}

\usepackage{amssymb}

\usepackage{capt-of}
\usepackage[hidelinks]{hyperref}

\begin{document}
%

\title{Open-Set Face Identification \\ on Few-Shot Gallery by Fine-Tuning}


\author{
\IEEEauthorblockN{Hojin Park, Jaewoo Park, and Andrew Beng Jin Teoh}
\IEEEauthorblockA{School of Electrical and Electronics Engineering\\
College of Engineering, Yonsei University\\
Seoul, Korea\\
2014142100@yonsei.ac.kr, julypraise@gmail.com, bjteoh@yonsei.ac.kr}
}

\maketitle

\begin{abstract}
In this paper, we focus on addressing the open-set face identification problem on a few-shot gallery by fine-tuning. The problem assumes a realistic scenario for face identification, where only a small number of face images is given for enrollment and any unknown identity must be rejected during identification. We observe that face recognition models pretrained on a large dataset and naively fine-tuned models perform poorly for this task. Motivated by this issue, we propose an effective fine-tuning scheme with classifier weight imprinting and exclusive BatchNorm layer tuning. For further improvement of rejection accuracy on unknown identities, we propose a novel matcher called Neighborhood Aware Cosine (NAC) that computes similarity based on neighborhood information. We validate the effectiveness of the proposed schemes thoroughly on large-scale face benchmarks across different convolutional neural network architectures. The source code for this project is available at: \href{https://github.com/1ho0jin1/OSFI-by-FineTuning}{https://github.com/1ho0jin1/OSFI-by-FineTuning}
\end{abstract}


%
\IEEEpeerreviewmaketitle

\section{Introduction}

\begin{figure}[t]
\centering
\subfloat[]{{\includegraphics[width=.48\textwidth]{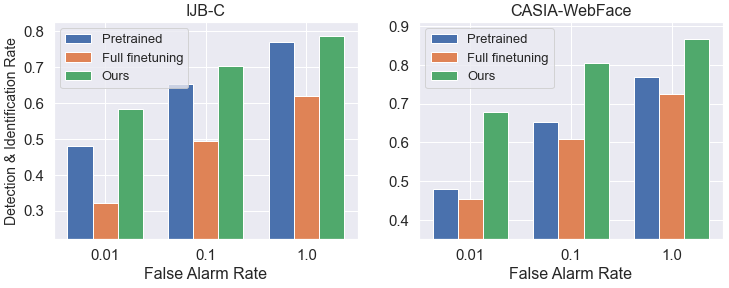} }\label{fig:mycaption-a}}
\\
\subfloat[]{{\includegraphics[width=.48\textwidth]{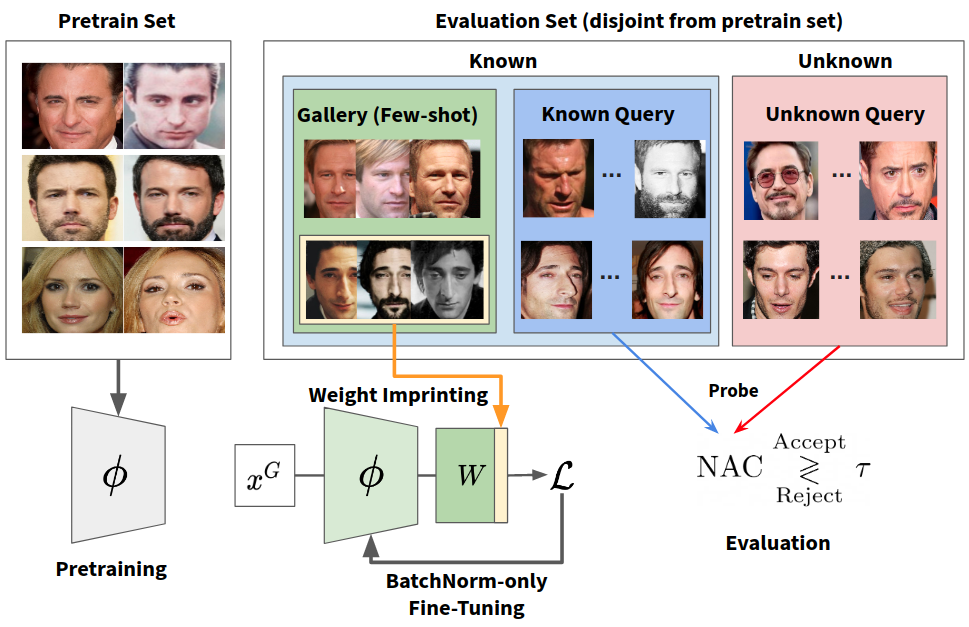} }\label{fig:mycaption-c}}
\caption{(a) Full fine-tuning all parameters severely degrades the OSFI performance, while our method significantly improves the pre-trained model. Detection \& Identification Rate (DIR) \cite{handbook} quantifies both correct identification of the known probe identities and detection of the unknown. (b) An outline of our proposed fine-tuning scheme: Given a model pretrained on a large-scale face database, we initialize the gallery set classifier by weight imprinting, and then fine-tune the model on a few-shot gallery set by training only the BatchNorm layers. In the evaluation stage, a given probe is either accepted as known or rejected as an unknown identity based on novel similarity matcher dubbed Neighborhood Aware Cosine (NAC) matcher.}
\label{fig:outline}
\end{figure}

Recently face recognition (FR) has achieved astonishing success attributed to three factors in large. Deep convolutional neural network (CNN) architectures \cite{VGGNet,ResNet} that have strong visual prior were developed and leveraged as FR embedding models. Large-scale face datasets \cite{cao2018vggface2,ms1m} that cover massive identities with diverse ethnicity and facial variations became available. On top of these, various metric learning losses \cite{schroff2015facenet,liu2017sphereface,wang2018cosface,deng2019arcface} elevated the performance of deep FR models to an unprecedented level.

The majority of FR embedding models have been evaluated on numerous benchmarks with closed-set identification \cite{liu2017sphereface,wang2018cosface,deng2019arcface,mvsoftmax,meng2021magface}. The closed-set identification protocol assumes all probe identities present in the gallery. However, in a realistic scenario, an unknown identity that is not enrolled may be encountered. Another important but practical aspect to consider is the scarcity of intra-class samples for the gallery identities to be registered; namely, due to the expensive data acquisition cost and privacy issue, only a very small number of samples might be available for each gallery identity to register. In this respect, \textit{open-set face identification} (OSFI) \textit{with the small-sized gallery} is closer to a real scenario as it needs to perform \textit{both} known probe identity identification and unknown probe identity rejection based on the limited information from the small gallery set. Despite its versatile practical significance, however, OSFI with a small gallery has been rarely explored.

Devising a model specific to OSFI with a small gallery can be challenging in the following aspects: Firstly, an OSFI model performs \textit{both} identifications of a known probe identity but also correct rejection of unknown probe identity. Hence, conventional FR embedding models devised mainly for closed-set identification can perform poorly at the rejection of the unknown.  In fact, as observed in Fig.~\ref{fig:outline} (a), FR embedding models pretrained on a large-scale public face database are not effective for open-set identification, leaving room for improvement. This suggests the need for fitting the pretrained model to be more specific to the given gallery set. 

Secondly, due to the \textit{few-shot} nature of the small-sized gallery set, there is a high risk of overfitting for fine-tuning the pretrained model. As shown in Fig.~\ref{fig:outline} (a), full fine-tuning (i.e. updating all parameters) of the pretrained model results in severe performance degradation. This drives us to devise an overfitting-resilient parameter tuning scheme.

Moreover, an ordinary cosine similarity matcher used in the closed-set identification might have a large tradeoff between the known probe identity identification and unknown probe identity rejection. As will be observed in Sec.~\ref{subsec:approach_nac}, the simple cosine matcher has a severe drawback for the task at hand. This motivates us to devise a robust matcher for OSFI.

Based on these observations, we propose an efficient fine-tuning scheme and a novel similarity-based matcher for OSFI constrained on a small gallery set. Our fine-tuning scheme consists of weight initialization of the classifier governed by weight imprinting (WI) \cite{imprint} and training only BatchNorm (BN) layers \cite{batchnorm} for overfitting-resilient adaptation on the small gallery set. Moreover, for both effective detection of the unknown and identification of the known probe identities, a novel Neighborhood Aware Cosine (NAC) matcher that respects the neighborhood information of the learned gallery features, and hence better calibrates the rejection score is proposed. Our contributions are summarized as follows:

\begin{enumerate}
\item To effectively solve the OSFI problem constrained on a small gallery set, we propose to fine-tune the pretrained face embedding model. Since full fine-tuning deteriorates the embedding quality, we search for the optimal method.
\item We demonstrate that the combination of weight imprinting and exclusive BatchNorm layer fine-tuning excels other baselines.
\item We recognize that the commonly used cosine similarity is a sub-optimal matcher for rejection. We propose a novel matcher named NAC that significantly improves the rejection accuracy.
\end{enumerate}

\section{Related Works}

\subsection{Open Set Face Identification (OSFI)}
\cite{transduction}, one of the earliest works in OSFI, used their proposed Open-set TCM-kNN on top of features extracted  by PCA and Fisher Linear Discriminant. \cite{osfr} proposed their own OSFI protocol and showed that an extreme value machine \cite{EVM} trained on the gallery set performs better than using cosine similarity or linear discriminant analysis for matchers. \cite{vareto2017towards} trained a matcher composed of locality sensitive hashing \cite{lsh_hash} and partial least squares\cite{mateos2011partial}. \cite{dao2021face} applied OpenMax \cite{openmax} and PROSER \cite{proser}, two methods for open-set recognition of generic images, on top of extracted face features.

All previous works propose to train an open-set classifier (matcher) of some form, but all of them use a fixed encoder. To the best of our knowledge, we are the first to propose an effective fine-tuning scheme as a solution to OSFI.\newline

\subsection{Cosine Similarity-based Loss Functions}
\cite{wang2017normface} proposed to $l_2$-normalize the features such that the train loss is only determined by the angle between the feature and the classifier weights. \cite{liu2017sphereface} further extended this idea by applying a multiplicative margin to the angle between a feature and its corresponding weight vector. This penalized the intra-class features to be gathered while forcing inter-class centers (prototypes) to be separated. A number of follow-up papers such as \cite{wang2018cosface,deng2019arcface,mvsoftmax,meng2021magface} modify this angular margin term in different ways, but their motivations and properties are generally similar. Therefore, in our experiments we only use CosFace loss \cite{wang2018cosface} as a representative method. For comprehensive understanding of these loss functions, refer to \cite{frsurvey}.

\section{Approach}
Our proposed approach is two-fold: fine-tuning on the gallery and open-set identification evaluation.
In the fine-tuning stage, the classifier is initialized by weight imprinting to initiate learning from optimal discriminative features, and the model is fine-tuned by updating only the BatchNorm layers to avoid overfitting on the few-shot gallery data. In evaluation, we utilize a novel matcher NAC that computes a neighborhood aware similarity for better-calibrated rejection of the unknown. We demonstrate that the combination of these three methods significantly outperforms all other baselines.

\subsection{Problem Definition and Metrics}
Formally, in an OSFI problem, we assume the availability of an encoder $\phi$ pretrained on a large-scale face database (FR embedding model), which is disjoint from the evaluation set with respect to identity. The evaluation set consists of a gallery $G = \{(x^G_i,y^G_i)\}^{Cm}_{i=1}$ and a probe set $Q$. The probe set $Q$ is further divided into the known probe set $K = \{(x^K_i,y^K_i)\}$ and the unknown probe set $U = \{(x^U_i,y^U_i)\}$. $G$ and $K$ has no overlapping images $x$ but shares same identities $y\in\{1,...,C\}$ whereas $U$ has disjoint identities, i.e., $\mathcal{Y}^U\cap\{1,...,C\}=\O$. $m$ refers to the number of images per identity in $G$, which we fix to 3 to satisfy the few-shot constraint. We allow the encoder to be fine-tuned over the gallery set.

The evaluation of OSFI performance uses the detection and identification rate at some false alarm rate (DIR@FAR). FAR=1 means we do not reject any probe.  Note that unlike the general case shown in \cite{handbook}, here we only consider rank-1 identification rate for DIR. Therefore, DIR@FAR=1 is the rank-1 closed-set identification accuracy.

\subsection{Classifier Initialization by Weight Imprinting}
\label{subsec:wi}
Due to the few-shot nature of the gallery set where we fine-tune on, the initialization of model parameters and, in particular, of classifier weights is crucial to avoid overfitting.
The most naive option is a random initialization of the classifier weight matrix $W$. Another commonly used strategy is linear probing \cite{anonymous2022finetuning}, i.e., finding an optimized weight $W$ that minimizes the classification loss over the frozen encoder embeddings $\phi(x)$.

\begin{figure}[t!]
\centering
\includegraphics[width=.5\textwidth]{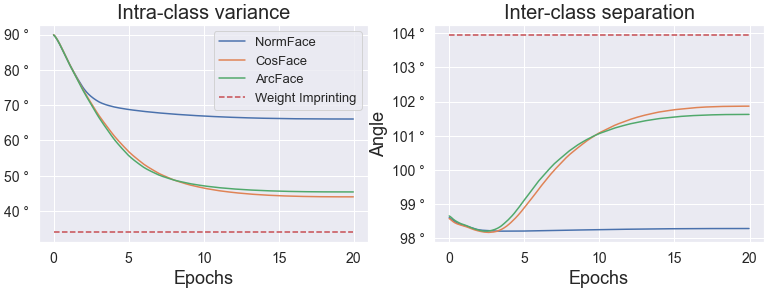}
\caption{The Intra-class variance (left) and inter-class separation  (right) of classifiers that are initialized by different schemes. NormFace \cite{wang2017normface}, CosFace \cite{wang2018cosface} and ArcFace \cite{deng2019arcface} loss are used for linear probing initialization. The weight imprinting initialization does not require training, thus stays constant.}
\label{fig:classifier_initialization}
\end{figure}

We experimentally find that, as seen in Fig.~\ref{fig:classifier_initialization}, both of these initialization schemes do not induce discriminative structure for the encoder embedding $\phi(x)$. In particular, during fine-tuning, each weight vector $w_c$ in the classifier acts as a center (or prototype) for the $c$-th class (i.e. identity). Fig.~\ref{fig:classifier_initialization} shows that neither random initialization nor linear probing of $w_c$ derives optimally discriminative weight vectors $w_c$, resulting in low quality of class separation of gallery features.

Motivated from this issue, we propose to initialize by \textit{weight imprinting} (WI), which induces the optimal discriminative quality for the gallery features:
\begin{equation}
w_c = \frac{\widehat{w_c}}{\lVert \widehat{w_c} \rVert_2},\quad \widehat{w_c} = \frac{1}{m}\sum_{y^G_i=c}\phi(x^G_i)
\end{equation}
where $\lVert \cdot \rVert_2$ is the $l_2$ norm, and the embedding feature $\phi(x)$ is unit-normalized such that $\lVert \phi(x) \rVert_2 = 1$.

As expected, Fig.~\ref{fig:classifier_initialization} verifies that fine-tuning from the weight imprinted initialization achieves a much higher discriminative quality. This shows the superiority of weight imprinting compared to random initialization and linear probing.

Note that weight imprinting has been frequently used in FR embedding models \cite{wang2018cosface,deng2019arcface}.  However, the critical difference is that those models utilize weight imprinting only to prepare templates before evaluation. In our case, the WI initialization is utilized particularly for fine-tuning.

\subsection{BatchNorm-only Fine-Tuning}
Choosing the appropriate layer to tune is another important issue for fine-tuning. 
Moreover, due to the extremely small number of samples for each gallery identity, there is a risk of overfitting as suggested by the classical theory on the vc dimension \cite{vapnik2015uniform}. In fact, a recent study \cite{anonymous2022finetuning} suggests that full fine-tuning hurts the pretrained filters including the useful convolutional filters learned from a large-scale  database. 

To minimize the negative effect of this deterioration, we fine-tune \textit{only the BatchNorm} (BN) \textit{layers} along with the classifier weight:
\begin{equation}
\min_{W,\,\theta_{BN}}\mathcal{L}(W^T \phi_\theta(x),y),
\quad \theta = [\theta_{BN},\theta_{rest}]
\end{equation}
where $\theta$ refers to all parameters in the encoder $\phi =\phi_\theta$ and $\theta_{BN}$ and $\theta_{rest}$ respectively refers to BatchNorm parameters and the rest. During fine-tuning, $\theta_{rest}$ is fixed with no gradient flow. The loss function $\mathcal{L}$ can be a softmax cross-entropy, or widely used FR embedding model losses such as ArcFace \cite{deng2019arcface} and CosFace \cite{wang2018cosface}.

Due to selective fine-tuning of only the BN layers (and classifier weight), the convolutional filters learned from the large-scale pre-train database are simply transferred. The BN-only training is thus computationally efficient as it occupies only 0.1-0.01\% of the total parameters in the CNN. Nevertheless, its model complexity is sufficient to learn a general image task as guaranteed by \cite{onlybn}.

\begin{minipage}[t]{.49\textwidth}
\begin{minipage}[b]{0.27\textwidth}
\centering
\includegraphics[width=\textwidth]{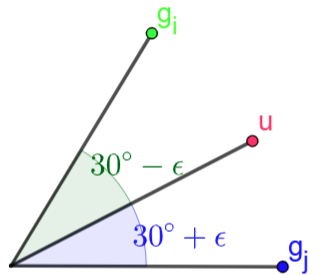}
\captionof{figure}{An unknown  feature $u$ placed between gallery prototypes of class $i$ and $j$. $\epsilon$ is some small positive constant.}
\label{fig:toy_example}
\end{minipage}
\begin{minipage}[b]{.7\textwidth}
\captionof{table}{Average angle (degrees) between IJB-C probe feature vectors and their top-k closest gallery prototypes. The third column refers to the average of top-2 to top-16.}
\label{table:known_unknown_angles}
\centering
\small
\begin{tabular}{ccccc}
\hline
Encoder                &                        & top-1 & top-2 & 2$\sim$16 \\ \hline
\multirow{2}{*}{Res50} & \multicolumn{1}{c|}{K} & 50.7$^{\circ}$  & 64.0$^{\circ}$  & 69.1$^{\circ}$                 \\
                      & \multicolumn{1}{c|}{U} & 63.8$^{\circ}$  & 66.0$^{\circ}$  & 69.7$^{\circ}$                 \\ \hline
\multirow{2}{*}{VGG19} & \multicolumn{1}{c|}{K} & 53.4$^{\circ}$  & 66.2$^{\circ}$  & 71.4$^{\circ}$                 \\
                      & \multicolumn{1}{c|}{U} & 65.9$^{\circ}$  & 68.2$^{\circ}$  & 72.1$^{\circ}$                 \\ \hline
\end{tabular}
\end{minipage}
\end{minipage}

\subsection{Neighborhood Aware Cosine Similarity}
\label{subsec:approach_nac}
The cosine similarity function is the most predominant matcher for contemporary face verification and identification. Denoting the probe feature vector as $p$ and the gallery prototypes as $\{g_j\}_{j=1}^C$, where $g_j := \frac{1}{m} \sum_{y^G_i = j} \phi(x^G_i)$ is the mean of all the normalized gallery feature vectors of class $j$, identification is performed by finding the maximum class index $c = \arg\max_{j=1,\dots,C}\cos(p,g_j)$. On the other hand, in the extension to OSFI, the decision of accepting as known or rejecting as unknown can be formulated:
\begin{equation}
\label{cosine_similarity}
\max_{j=1,\dots,C}\cos(p,g_j)\underset{\text{Reject}}{\overset{\text{Accept}}{\gtrless}}\tau
\end{equation}
where $\cos(p,q) = \frac{p}{\lVert p \rVert_2} \cdot \frac{q}{\lVert q \rVert_2}$ is the cosine similarity between two feature vectors, $\tau$ is the rejection threshold.

Now, consider an example illustrated in Fig.~\ref{fig:toy_example}. The cosine matcher will assign the probe $u$ to the identity $i$ with the acceptance score $0.866$, which is fairly close to the maximum score $1$. This value alone might imply that the probe is a known sample as it is close to the gallery identity $i$. However, the probe feature vector is placed right in the middle of the identities $i$ and $j$. The in-between placement of $u$ suggests that the probe can be possibly unknown and thus should be assigned with a lesser value of the acceptance score.

\begin{figure}[t]
\centering
\includegraphics[width=.42\textwidth]{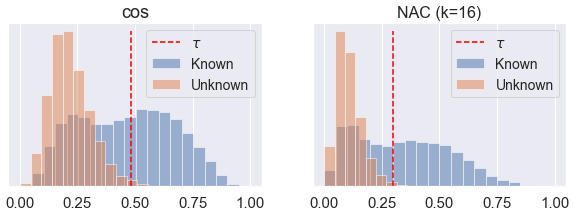}
\caption{The distributions of scores for known (K) and unknown (U) probes of IJB-C dataset using cosine similarity (left) and NAC with $k=16$ (right). The scores are min-max normalized and $\tau$ is set such that FAR=0.01 for both cases. DIR=48.05\% (left) vs DIR=54.53\% (right). ResNet-50 was used as the encoder.}
\label{cos vs nac}
\end{figure}

Motivated by this intuition, we propose the \textit{Neighborhood Aware Cosine} (NAC) matcher that respects all top-$k$ surrounding gallery features:
\begin{equation}
\label{NAC}
\text{NAC}(p,g_i) = \frac{\exp(cos(p,g_i))\cdot\text{1}[i \in N_k]}{\sum_{j\in N_k}\exp(\cos(p,g_j))}
\end{equation}
Here, $N_k$ is the index set of $k$ gallery prototypes that are nearest to the probe feature $p$, and $\text{1}$ is the indicator function. The main goal of the NAC matcher is to improve the unknown rejection. Table \ref{table:known_unknown_angles} shows that known probe features are much closer to their closest prototype than the second-closest prototype, unlike unknown probes. By exploiting this phenomenon, the NAC matcher is able to assign a much smaller score to unknown probe, as shown in Fig.~\ref{cos vs nac}.

\section{Experiments}
\subsection{Datasets}
We use VGGFace2 \cite{cao2018vggface2} dataset for pretraining the encoders, and CASIA-WebFace \cite{CASIA} and IJB-C \cite{IJBC} for evaluation. Using MTCNN \cite{mtcnn}, we align and crop every images to 112x112 with equal parameters for all datasets. For VGGFace2, we remove all identities overlapping with the evaluation datasets. The evaluation datasets are equally split into two groups such that the number of known and unknown identities are equal. Then we randomly choose $m{=}3$ images of the known identities to create the gallery (G), and the rest are known probes (K). All images of unknown identities are unknown probes (U). Table \ref{dataset statistics} summarizes the statistics of the datasets we use. Note that we chose every known identity to have more than 10 images such that there can be at least 7 probe samples. Also note that IJB-C dataset consists of still images and video frames (video frames typically have poorer image quality). We sample the gallery from still images and probes from video frames, which makes this dataset much challenging. We note that the protocol devised here can be regarded as an extension of that in \cite{osfr}.

\subsection{Baselines}
\label{subsec:baselines}
\subsubsection{Classifier Initialization}
Along with Weight Imprinting (denoted \textbf{WI}), we report the results of using \textbf{random initialization} and \textbf{linear probing} initialization as described in Sec.~\ref{subsec:wi}.

\subsubsection{Encoder Layer Fine-Tuning}
Along with BatchNorm-only fine-tuning (denoted as \textbf{BN}), we explore tuning other layers of the encoder. The simplest one is tuning every layer (i.e. all parameters of a model), which we denote as \textbf{full}. The second is freezing the early layers and training only the deeper ones, which we denote as \textbf{partial}. We also consider the parallel residual adapter \cite{parallel}, which adds additional 1x1 convolutional filters to the original convolutional layers. During fine-tuning, only these additional filters are trained to capture the subtle difference in the new dataset. Note that the authors in \cite{parallel} apply this technique to ResNet \cite{ResNet}, hence the name \textit{residual} parallel adapter. But this idea can be generally applied to CNNs without residual connection, hence we also apply this to a VGG-style network. We denote this as \textbf{PA}, referring to Parallel Adapter.

\begin{table}[t!]
\setlength\doublerulesep{2mm}
\centering
\caption{Dataset Statistics. The number inside the parentheses refers to the average number of images per identity. For evaluation datasets, Known identities consist of the gallery (G) and known probe (K), where the gallery has 3 images per identity.}
\label{dataset statistics}
\begin{tabular}{|c|cc|}
\hline
Pretrain      & \multicolumn{2}{c|}{\# IDs (images / ID)}            \\ \hline
VGGFace2      & \multicolumn{2}{c|}{7,689 (354.0)}                 \\ \hline\hline
Evaluation    & \multicolumn{1}{c}{Known (G + K)}  & Unknown (U)  \\ \hline
CASIA-WebFace & \multicolumn{1}{c}{5,287 (3+20.0)} & 5,288 (16.5) \\
IJB-C         & \multicolumn{1}{c}{1,765 (3+15.3)} & 1,765 (13.9) \\ \hline
\end{tabular}
\end{table}

\begin{table}[t]
\centering
\caption{The total number of parameters and number of fine-tuned parameters for each encoder fine-tuning scheme. `+' refers to the number of added parameters for the Parallel Adapter.}
\label{encoder configs}
\begin{tabular}{c|rr}
\hline
~ & \multicolumn{2}{c}{\# Params (million)} \\ \cline{2-3}
~ & VGG19 & Res50 \\ \cline{2-3}
Pretrained & 32.88 & 43.58 \\ \hline
Full fine-tuning & 32.88 & 43.58 \\
Partial fine-tuning & 4.72 & 4.72 \\
Parallel Adapter & +2.22 & +3.39 \\
BN-only fine-tuning & 0.01 & 0.03 \\ \hline
\end{tabular}
\end{table}

\subsubsection{Matcher}
During OSFI evaluation, the vanilla \textbf{cos}ine similarity matcher is adopted as the baseline matcher. When the NAC matcher is used, we denote by \textbf{NAC}. For comparison, we also use the extreme value machine (\textbf{EVM}) proposed by \cite{osfr}. We train the EVM on the gallery set with the best parameters found by the authors.

In summary, classifier initialization methods we consider are \{\textbf{Random}, \textbf{Linear probing}, \textbf{WI}\}, fine-tuning layer configurations are \{\textbf{Full}, \textbf{Partial}, \textbf{PA}, \textbf{BN}\}, and matchers are \{\textbf{cos}, \textbf{EVM}, \textbf{NAC}\}. We test the OSFI performances among different combinations of these three components. Our proposed OSFI scheme is to use \textbf{WI+BN+NAC} jointly.

\begin{figure*}[t]
\centering
\includegraphics[width=0.9\textwidth]{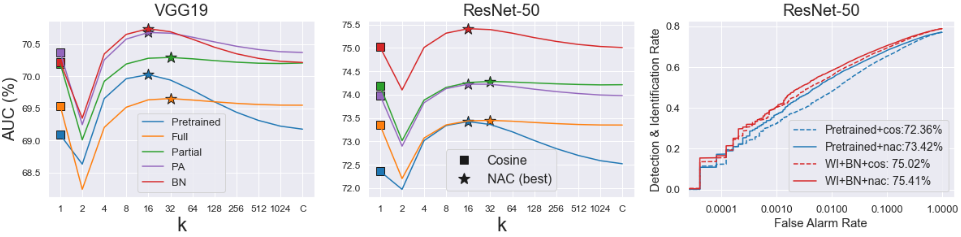}
\caption{The OSFI performance of cosine similarity and NAC with different values of $k$ on IJB-C dataset, using VGGNet-19 (left) and ResNet-50 (mid) as the encoder. The square markers refer to cosine similarity and star marks the optimal \textit{k} for different layer fine-tuning methods. To summarize the OSFI performance into a single number, we used the area under the curve (AUC, \%) of DIR@FAR curve. (Right) DIR@FAR curve of \textit{Pretrained} and \textit{BN} configuration using cosine similarity and NAC (\textit{k}=16) as the matcher. Numbers in the legend show the AUC values. When $k=1$, NAC is replaced by $\cos$.}
\label{NAC best k}
\end{figure*}

\subsection{Training Details}
We choose VGG19 \cite{VGGNet} and ResNet-50 \cite{ResNet} for the encoders with the feature dimension 512. We pretrain these encoders on the VGGFace2 dataset with CosFace with scale=32, margin=0.4 as loss function until convergence.

Then we fine-tune the encoder with different classifier initialization schemes and encoder layer configurations. When using the linear probing initialization, we train the classifier until the training accuracy reaches 95\%.

We follow the encoder layer finetuning in Sec.~\ref{subsec:baselines}. For the partial fine-tuning, we only train the last 2 convolutional layers (Conv-BN-ReLU-Conv-BN-ReLU). Table \ref{encoder configs} shows the number of total and updated parameters for each fine-tuning scheme.

We fix the number of epochs to 20 and batch size to 128 for every method. We again use CosFace loss for consistency. For the optimizer we use Adam \cite{adam} with cosine annealing. The initial learning rate is set to 1e-4 for \textit{full} and \textit{PA}, and 1e-3 for \textit{partial} and \textit{BN}, which we find as the optimal learning rate for each method. For data augmentation, we use random horizontal flipping and random cropping with the random scale from 0.7 to 1.0. The cropped images are resized to the original size.

\begin{table*}[]
\centering
\caption{DIR@FAR of different methods on CASIA-WebFace dataset and IJB-C dataset, using VGGNet-19 and ResNet-50 as the encoder. DIR@FAR=1 (100\%) is the closed-set accuracy. The highest value in each column is marked in bold. For the first three rows the encoder is not fine-tuned and only the matchers are changed. The last row (\textbf{WI+BN+NAC}) is our proposed method.}
\label{all results}
\begin{tabular}{c|lll|cccc|cccc}
\hline
\multirow{3}{*}{Encoder} & \multicolumn{3}{c|}{Method}                                                                                                                                                              & \multicolumn{4}{c|}{CASIA-WebFace}                                & \multicolumn{4}{c}{IJB-C}                                        \\ \cline{2-12} 
                         & \multirow{2}{*}{\begin{tabular}[c]{@{}l@{}}Classifier\\ initialization\end{tabular}} & \multirow{2}{*}{\begin{tabular}[c]{@{}l@{}} Fine-tuning\\ layers\end{tabular}} & \multirow{2}{*}{Matcher} & \multicolumn{4}{c|}{DIR @ FAR (\%)}                                    & \multicolumn{4}{c}{DIR @ FAR (\%)}                                    \\
\cline{5-12}
                         &                                                                                      &                                                                        &                          & 0.1          & 1.0           & 10.0            & 100.0            & 0.1          & 1.0           & 10.0            & 100.0            \\ \hline
\multirow{10}{*}{VGG19}   & None                                                                                 & None                                                                   & cos                      & 25.23          & 52.97          & 70.07          & 80.89          & 28.35          & 45.55          & 61.71          & 73.80          \\
                         & None                                                                                 & None                                                                          & EVM                      & \textbf{37.57} & 57.75          & 71.03          & 80.78          & 35.03          & \textbf{53.64} & 63.34         & 73.70           \\
                         & None                                                                                 & None                                                                          & NAC                      & 25.15          & 55.68          & 71.41          & 80.89          & 36.73          & 51.92          & 64.27         & 73.80           \\
                         & Random                                                                               & Full                                                                   & cos                      & 23.95          & 43.19          & 59.03          & 70.94          & 17.18          & 32.62          & 46.90          & 60.23          \\
                         & Linear probing                                                                       & Full                                                                   & cos                      & 28.82          & 55.64          & 70.44          & 79.84          & 30.80          & 45.91          & 59.63          & 70.09          \\
                         & WI                                                                                   & Full                                                                   & cos                      & 27.63          & 57.58          & 72.02          & 80.94          & 35.49          & 50.52          & 63.56          & 73.53          \\
                         & WI                                                                                   & Partial                                                                & cos                      & 28.91 & 57.31          & 72.29          & 81.16          & 34.81          & 51.98          & 64.53          & 73.89          \\
                         & WI                                                                                   & PA                                                                     & cos                      & 26.29          & 57.90          & 72.82          & 81.82          & 31.74          & 50.21          & 64.26          & \textbf{74.50} \\
                         & WI                                                                                   & BN                                                                     & cos                      & 25.39          & 56.65          & 72.54          & 82.14          & 32.19          & 48.74          & 63.87          & 74.43          \\
                         & WI                                                                                   & BN                                                                     & NAC                      & 25.94          & \textbf{58.01} & \textbf{72.92} & \textbf{82.14} & \textbf{38.09} & 53.08 & \textbf{65.30} & 74.43          \\ \hline
\multirow{10}{*}{Res50}   & None                                                                                 & None                                                                   & cos                      & 23.85          & 58.06          & 74.15          & 83.69          & 32.11          & 48.05          & 65.31          & 76.96          \\
                         & None                                                                                 & None                                                                          & EVM                      & \textbf{39.44} & 61.61          & 75.02          & 83.57          & 38.12          & 38.12          & 66.81         & 76.96          \\
                         & None                                                                                 & None                                                                          & NAC                      & 21.24          & 60.23          & 75.31          & 83.69          & 36.67          & 54.53          & 68.14         & 76.96          \\
                         & Random                                                                               & Full                                                                   & cos                      & 25.31          & 45.43          & 60.80          & 72.44          & 14.88          & 32.05          & 49.39          & 61.88          \\
                         & Linear probing                                                                       & Full                                                                   & cos                      & 28.35 & 60.11          & 74.63          & 82.73          & 30.35          & 46.42          & 61.90          & 72.34          \\
                         & WI                                                                                   & Full                                                                   & cos                      & 26.73          & 63.92          & 77.49          & 84.65          & 39.05          & 56.00          & 67.83          & 76.94          \\
                         & WI                                                                                   & Partial                                                                & cos                      & 25.98          & 64.66          & 78.07          & 85.02          & \textbf{44.31} & 57.11          & 69.13          & 77.49          \\
                         & WI                                                                                   & PA                                                                     & cos                      & 24.89          & 63.85          & 77.58          & 85.01          & 36.69          & 54.86          & 68.30          & 77.63          \\
                         & WI                                                                                   & BN                                                                     & cos                      & 25.70          & 65.83          & 79.66          & 86.73          & 40.29          & 55.71          & 69.29          & 78.74          \\
                         & WI                                                                                   & BN                                                                     & NAC                      & 23.65          & \textbf{67.72} & \textbf{80.34} & \textbf{86.73} & 40.25          & \textbf{58.25} & \textbf{70.40} & \textbf{78.74} \\ \hline
\end{tabular}
\end{table*}

\subsection{Optimal \textit{k} for NAC}
Since the gallery set is too small, we cannot afford a separate validation set to individually optimize $k$ for each dataset. Instead, we attempt to find a global value that has optimal performance regardless of the fine-tuning method, if one exists.

We first fine-tune the encoders with different layer configurations, which gives us five different encoders including one without any fine-tuning; \textit{pretrained, full, partial, PA}, and \textit{BN}. Then we search the best parameter \textit{k} for the NAC matcher by grid search strategy, where the grid is [2,4,8,16,32,128,256,512,1024,$C$], and $C$ is the total number of identities. Note that $k=1$ refers to using cosine similarity instead of NAC, which we added for comparison. Since a single-value objective is preferred, we use the area under the curve (AUC) of the DIR@FAR curve instead of DIR value at different FAR values. We repeat this process with different datasets and encoder architectures.

The results are shown in Fig. \ref{NAC best k}. We did not include the results of CASIA-WebFace as it shows a similar trend. Excluding $k=1$ which is not NAC, the results show a smooth unimodal curve with a peak at $k=16$ or $32$. This shows that the NAC matcher indeed has a globally optimal $k$ value that is robust against different datasets, encoders, and fine-tune methods. Thus we choose $k=16$ ($k=32$ also gives similar results) as the global parameter throughout this paper.

Note that when $k=C$, NAC becomes equivalent to softmax function with cosine similarity logits. However, this is notably inferior compared to $k=16$, which implies that considering only the $k$-nearest is superior to considering \textit{every} gallery prototype.

\subsection{Comparison of Fine-Tuning Methods}
We compare the OSFI performances of the pretrained model (non-fine-tuned) with six different combinations of classifier initialization schemes and layer finetuning configurations: random+full, linear probing+full, WI+full, WI+partial, WI+PA, WI+BN. The matcher is fixed to cosine similarity. These correspond to row 4-9 in Table \ref{all results}.

First, to compare different classifier initialization schemes, we fix the fine-tuning scheme to \textit{full}. When using \textit{random} initialization, rejection accuracy (DIR@FAR=0.001,0.01,0.1) and closed-set accuracy (DIR@FAR=1) severely drops. For \textit{linear probing}, rejection accuracy improves while closed-set accuracy drops. Only \textit{WI} clearly improves the encoder performance, supporting the superiority of weight imprinting.

Now we fix the classifier initialization to \textit{WI} and compare different layer finetuning configurations. \textit{full} clearly has the worst performance. While \textit{PA} is better than \textit{partial} in closed-set accuracy, \textit{partial} clearly outperforms \textit{PA} in rejection accuracy. \textit{BN} outperforms all others in closed-set accuracy with a large margin but sometimes falls behind \textit{partial} in rejection accuracy.

With the aid of the NAC matcher, our method WI+BN+NAC outperforms all other methods in every aspect. Compared to \textit{original}, this gains 4.60\%, 8.11\%, 4.57\%, 1.68\% higher DIR  in average with respect to FAR of 0.001, 0.01, 0.1, 1.0, respectively.

\begin{table}[t!]
    \centering
    \caption{Inter-class separation, intra-class variance, DBI, and AUC gain by using NAC (refer to Fig. \ref{NAC best k}) for each layer finetuning configuration. These values are averaged across datasets and encoder architectures. $\uparrow$ means that larger quantity is better and vice versa.}
    \begin{tabular}{c|cccc}
    \hline
        ~ & Inter ($\uparrow$) & Intra ($\downarrow$) & DBI ($\downarrow$) & $\Delta$AUC ($\uparrow$) \\ \hline
        Pretrained Model & 106.3$^{\circ}$ & 34.5$^{\circ}$ & 1.52 & 0.740 \\
        Full finetuning & 106.7$^{\circ}$ & 24.2$^{\circ}$ & 0.87 & 0.025 \\
        Partial finetuning & 106.4$^{\circ}$ & 24.5$^{\circ}$ & 0.90 & 0.058 \\
        Parallel Adapter & 107.0$^{\circ}$ & 31.8$^{\circ}$ & 1.32 & 0.135 \\
        BN-only finetuning & 107.3$^{\circ}$ & 33.6$^{\circ}$ & 1.46 & 0.335 \\ \hline
    \end{tabular}
\label{inter intra dbi}
\vspace{-5mm}
\end{table}

\subsection{Analysis on Discriminative Quality of Different Fine-tuning Methods}
How do different layer finetuning configurations affect the final OSFI performance? To analyze this, we adopt three different metrics; inter-class separation, intra-class variance, and Davies-Bouldin Index (DBI) \cite{dbi}. The definitions of the first two metrics are identical to that of Fig.~\ref{fig:classifier_initialization}. DBI is a metric for evaluating the clustering quality, where DBI $\approx 0$ means perfect clustering. We compute these metrics on the gallery features after fine-tuning, and the results are shown in Table \ref{inter intra dbi}.

Here we can easily separate these configurations into two groups: \textit{full} and \textit{partial} vs \textit{PA} and \textit{BN}. The first group has similar inter-class separation with \textit{Pretrained} and significantly smaller intra-class variance, which leads to small DBI. This is in stark contrast with the second group.

With this observation, we can conjecture the different optimization strategies of each group. The first group was able to easily reduce the training loss by \textbf{collapsing} the gallery features into a single direction (shown by the small angle between intra-class features). This was possible because both \textit{full} and \textit{partial} directly updated the parameters of the convolutional filters. On the other hand, all convolutional filters were frozen for both \textit{PA} and \textit{BN}. This constraint may have prevented these methods from taking the shortcut, i.e. simply collapsing the gallery features, and instead led to separating the embeddings of different identities. This explains why \textit{PA} and \textit{BN} have higher closed-set accuracy.

This can also explain the AUC gain ($\Delta$AUC) when using NAC instead of cosine similarity. Features become redundant when they collapse, and so does the prototype. Therefore the information from neighboring prototypes becomes less helpful in rejecting unknown samples, leading to the marginal gain from using NAC. This is why \textit{full} and \textit{partial} do not benefit from using NAC matcher.

\begin{figure}
    \centering
    \includegraphics[width=0.5\textwidth]{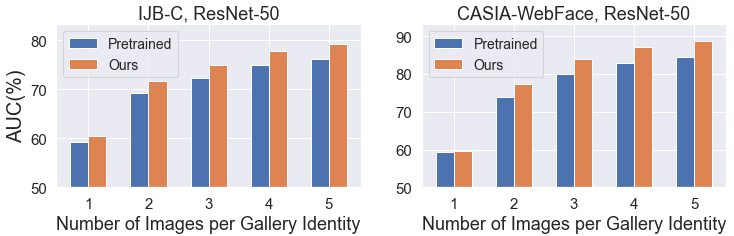}
    \caption{The performance of our method against the baseline w.r.t. different gallery size. AUC of DIR@FAR curve is used as the performance measure.}
    \label{fig:num_images}
\end{figure}

\subsection{Performance with respect to Different Gallery Size}
Fig.~\ref{fig:num_images} shows the OSFI performance of our method against the baseline (pretrained encoder with \textbf{cos} matcher) with respect to different gallery size. We can see that our method consistently improves upon the baseline, except for the extreme case where only one image is provided for each identity.

\section{Conclusion and Future Works}
In this work we showed that combining weight-imprinted classifier and BatchNorm-only tuning of the encoder effectively improves the encoder's OSFI performance without suffering from overfitting. We further facilitated the performance by our novel NAC matcher instead of the commonly used cosine similarity.
Future works will explore extending this idea to the open-set few-shot recognition of generic images.\newline
\textbf{Acknowledgements:}\newline
This work was supported by the National Research Foundation of Korea (NRF) grant funded by the Korea government (MSIP) (NO. NRF-2022R1A2C1010710)

\bibliographystyle{IEEEtran}
\bibliography{IEEEexample}

\end{document}